\documentclass[a4paper,fleqn]{cas-dc}

\usepackage[numbers,sort&compress]{natbib}
\usepackage{amsmath}
\usepackage[dvipsnames]{xcolor}
\usepackage{soul}
\setstcolor{RoyalBlue}

\def\tsc#1{\csdef{#1}{\textsc{\lowercase{#1}}\xspace}}
\tsc{WGM}
\tsc{QE}

\begin{document}
\let\WriteBookmarks\relax
\def\floatpagepagefraction{1}
\def\textpagefraction{.001}

\shorttitle{Diffusion Unit}    

\shortauthors{Xiu et al.}  

\title [mode = title]{Diffusion Unit: Interpretable Edge Enhancement and Suppression Learning for 3D Point Cloud Segmentation}  



%

\author[1,2]{Haoyi Xiu}
\author[1]{Xin Liu}
\cormark[1]
\author[1]{Weimin Wang}
\fnmark[1]
\fntext[1]{Also is affiliated with Dalian University of Technology}
\author[1]{Kyoung-Sook Kim}
\author[4]{Takayuki Shinohara}
\author[5]{Qiong Chang}
\author[2]{Masashi Matsuoka}





\affiliation[1]{organization={Artificial Intelligence Research Center, AIST},
            city={Tokyo},
            country={Japan}}
            
\affiliation[2]{organization={Department of Architecture and Building Engineering, Tokyo Institute of Technology},
            city={Tokyo},
            country={Japan}}


\affiliation[3]{organization={Innovation Technology Office Research Center, PASCO Corporation},
            city={Tokyo},
            country={Japan}}

\affiliation[4]{organization={Department of Computer Science, Tokyo Institute of Technology},
            city={Tokyo},
            country={Japan}}







\cortext[1]{Corresponding author.\\E-mail: xin.liu@aist.go.jp (Xin Liu) \\wangweimin@dlut.edu.cn (Weimin Wang)}

















\begin{abstract}
3D point clouds are discrete samples of continuous surfaces which can be used for various applications. However, the lack of true connectivity information, i.e., edge information, makes point cloud recognition challenging.
Recent edge-aware methods incorporate edge modeling into network designs to better describe local structures.
Although these methods show that incorporating edge information is beneficial, \textit{how} edge information helps remains unclear, making it difficult for users to analyze its usefulness.   
To shed light on this issue, in this study, we propose a new algorithm called Diffusion Unit (DU) that handles edge information in a principled and interpretable manner while providing decent improvement. 
First, we theoretically show that DU learns to perform task-beneficial edge enhancement and suppression. 
Second, we experimentally observe and verify the edge enhancement and suppression behavior. 
Third, we empirically demonstrate that this behavior contributes to performance improvement. 
Extensive experiments and analyses performed on challenging benchmarks verify the effectiveness of DU. Specifically, our method achieves state-of-the-art performance in object part segmentation using ShapeNet part and scene segmentation using S3DIS. Our source code is available at \url{https://github.com/martianxiu/DiffusionUnit}.
\end{abstract}




\maketitle

\section{Introduction}\label{sec: introduction}
A 3D point cloud is a basic and flexible shape representation where the surface is represented as a set of discrete points in the 3D space. Powered by the recent advancement of cost-effective sensor technology, numerous large-scale datasets are publicly available to research communities, facilitating deep learning--based point cloud understanding. The application field of such research can be autonomous driving~\cite{qi2018frustum,cui2021deep, wen2022deep}, remote sensing~\cite{shinohara2020fwnet,xiu2023ds}, and so on.

Among diverse types of deep neural networks (DNNs), Convolutional Neural Network (CNN) is the most effective algorithm for computer vision. However, the unstructured and unordered nature of point clouds prevents researchers from directly applying CNN to point clouds. 
Early works often project point clouds to regular grids before applying CNN \cite{kanezaki2018rotationnet,su2015multi,maturana2015voxnet,zhou2018voxelnet} or utilizing special data structures~\cite{wang2017cnn,tatarchenko2017octree,klokov2017escape}.
However, most of these methods may suffer from information loss caused by projections. This issue is resolved by PointNet~\cite{qi2017pointnet}, which consists of permutation-invariant operations and can handle point clouds in a lossless manner. Further, PointNet++ extends PointNet to achieve hierarchical learning~\cite{qi2017pointnet++}. Based on the PointNet++ framework, varieties of convolution methods~\cite{li2018pointcnn,wu2019pointconv,thomas2019kpconv,xu2021paconv,qian2022pointnext} and transformers~\cite{yang2019modeling,zhao2021point,lai2022stratified,wu2022pointv2,guo2022ctpoint} for point clouds are proposed. 

On the other hand, point clouds naturally lack structures because they are discrete point samples of continuous surfaces. This means that true connectivity information, i.e., edge information, is not available. Hence, incorporating edge modeling into model designs can help to learn local structures of continuous surfaces, which in turn leads to better performance in varieties of high-level tasks~\cite{wang2019dynamic,liu2020closer,xu2021paconv}, especially in those that benefit from optimized connectivity.
Edge information, or the edge feature, is often represented by the output of neural network functions (e.g., multi-layer perceptions (MLPs)) that take features of a center point and its neighbor as input~\cite{wang2019dynamic}.
The resulting edge features are used, for instance, as extra descriptors describing the local structure~\cite{liu2020closer,xiang2021walk} or as similarity measures reflecting spatial consistency~\cite{wang2019graph,zhao2021point}.
Although these edge-aware methods empirically show that the incorporation of edge features improves performance, how they work to improve performance remains unclear.
This lack of interpretability can be problematic because 1) it makes it difficult for users to identify fruitful research directions when it works well and 2) it also makes it difficult for users to diagnose the problem when they fail. 

In this study, we propose the \textit{Diffusion Unit (DU)} that handles edges in a principled and interpretable manner while providing decent performance gain. 
DU extends the classic diffusion theory~\cite{perona1990scale,weickert1998anisotropic} that is often used for edge-preserving smoothing. An important motivation for extending the diffusion theory is that it provides a mathematically transparent framework that makes the behavior of DU interpretable. 
Specifically, we theoretically show that DU learns to enhance task-beneficial edges (e.g., part boundaries of objects) and suppress unhelpful discontinuities. 
Furthermore, we experimentally observe and verify the edge enhancement and suppression behavior of DU through intuitive visualizations. 
Moreover, we empirically demonstrate that this behavior contributes to performance improvement. 

\begin{figure}[t]
    \centering 
        \includegraphics[width=0.8\linewidth]{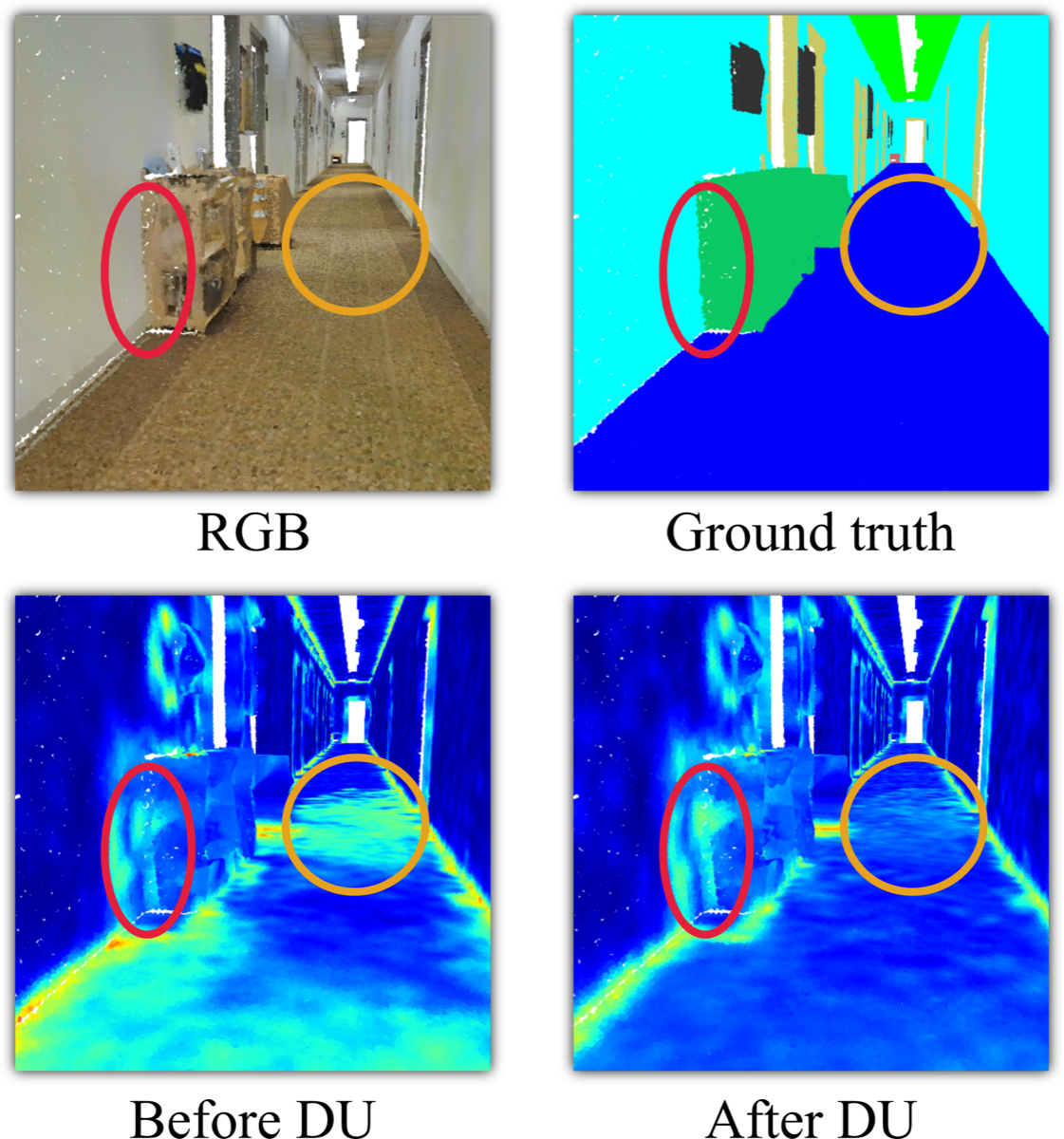}
    \caption{The effect of DU. The heat map indicates the local feature smoothness. A low response indicates a smooth region while a high response indicates a non-smooth region. Red and orange circles indicate enhanced and suppressed features after applying DU. As can be seen, task-beneficial edges (in this case, object boundaries) are enhanced/preserved whereas intra-region discontinuities are suppressed.  
    }
    \label{fig: concept}
\end{figure}

To validate the interpretability and effect of DU, the task used must simultaneously meet two conditions: 1) interpretability can be validated. That is, the task has clear definitions of task and non-task edges so that we can empirically verify the edge enhancement and suppression behavior of DU; 2) Modeling the connectivity/edges between points is important. That is, the optimization of edges must be closely related to the task objective so that the effect of DU can be fairly and confidently evaluated. In this study, we adopt part and scene segmentation tasks for the validation of DU, as they clearly satisfy the above two conditions.
Therefore, we construct \textit{DU-Net}, a network specialized for point cloud segmentation, with DU as its core building block.
Extensive experiments are performed to verify the edge enhancement/suppression behavior of DU as well as the practical effectiveness of DU-Net. In particular, we achieve state-of-the-art performance in object part segmentation using ShapeNet part~\cite{yi2016scalable} and scene segmentation using S3DIS~\cite{armeni20163d}. The effect of DU is shown in Fig.~\ref{fig: concept}.

Our main contributions are summarized as follows:
\begin{itemize}
    \item We propose DU that performs edge enhancement and suppression learning in a principled manner.
    \item We theoretically analyze and experimentally demonstrate the edge enhancement and suppression behavior of DU.  
    \item We design DU-Net for point cloud segmentation and perform extensive experiments to verify its practical effectiveness and design choices. 
\end{itemize} 

The remainder of this paper is organized as follows. Section~\ref{sec:related_work} revisits the related works concerning deep learning methods for 3D point clouds. In particular, we provide a detailed review of edge-aware methods that use edge features to better model local structures. Since diffusion theory is often used for edge-preserving smoothing and our work is closely related to it, diffusion-related methods are subsequently reviewed to show how it is developed and applied to various tasks. 
Section~\ref{sec:method} first elaborates on the motivation to introduce explicit edge modeling. Next, we introduce the diffusion equation which forms the basis of DU. Then, we explain the challenges of applying the diffusion equation directly to point clouds and propose DU in its continuous form. We subsequently conduct a theoretical analysis and show that DU is guaranteed to perform adaptive edge enhancement and suppression. To apply DU to discrete point clouds, we introduce discretization schemes. Finally, we construct DU-Net for point cloud segmentation with  DU as its core component.    
Section~\ref{sec: exp} empirically verifies the theoretical results of Section~\ref{sec:method} through visualizations. Moreover, experiment results on part and scene segmentation tasks are also reported to validate its practical effectiveness. Then, ablation studies are also made to validate our design choices.
We conclude this study in Section~\ref{sec:conclusion}.

\section{Related Work}
\label{sec:related_work}
In this section, we first provide a brief review of deep learning methods for point cloud analysis. Then, we review approaches that improve point cloud understanding with the explicit awareness of edge information. As the diffusion equation is often used for edge-preserving smoothing and we also use it as DU's basis, we subsequently provide a brief review of methods that use diffusion equations.   

\subsection{Deep learning for 3D point clouds}
\paragraph{Projection-based methods} 
Owing to the unstructured and unordered nature of point clouds, early methods that deal with point clouds are projection-based methods. Those methods project point clouds to regular grids (e.g., 2D pixels~\cite{kanezaki2018rotationnet,su2015multi,feng2018gvcnn} and 3D voxels~\cite{maturana2015voxnet,wang2017cnn,tatarchenko2017octree,zhou2018voxelnet,graham20183d,choy20194d}) to enable matured regular convolution to operate on point clouds. The problem with these methods is that, however, they lose fine-grained geometric details through projections. 

\paragraph{MLP-based methods}
MLP-based methods operate directly on the raw point clouds, thereby preventing information loss due to projections. The representative method is PointNet~\cite{qi2017pointnet} whose main operations consist of shared MLPs and symmetric functions (e.g., max-pooling). Both operations are permutation-invariant, and thus the irregularity of point clouds is well-resolved. To further enhance PointNet, PointNet++~\cite{qi2017pointnet++} applies PointNets to local subsets of points grouped according to different scales or resolutions to better model the local geometric details. By repeating such a process multiple times with increasing grouping scales, PointNet++ achieves efficient CNN-like hierarchical learning. Afterward, various advancements are made based on PointNet and PointNet++~\cite{lan2019modeling,qiu2021geometric}. For instance, ShellNet~\cite{zhang2019shellnet} further divides a local neighborhood into multiple groups and applies MLPs and symmetric functions to all groups, aiming to leverage fine-grained details. PointNeXt~\cite{qian2022pointnext}, on the other hand, constructs a modern version of PointNet++ by carefully combining the advanced training and scaling strategies. RepSurf~\cite{ran2022surface} uses an explicit representation of surfaces based on several geometric priors and successfully enhances PointNet++. PointMixer~\cite{choe2022pointmixer} which is inspired from MLP-Mixer~\cite{tolstikhin2021mlp} is proposed. It realizes intra-set, inter-set, and hierarchical-set mixing to adapt the original mixing operation to point cloud processing.

\paragraph{Point convolution--based methods}
The point convolution--based method defines a convolution operation on unstructured point clouds. For instance, some studies construct a pseudo grid for each convolution local neighborhood on which input points are projected~\cite{thomas2019kpconv,mao2019interpolated, liu2020closer}. Since the constructed grid is structured, a standard convolution operation becomes applicable to projected point features. On the other hand, some methods predict convolution filters on the fly~\cite {li2018pointcnn,wu2019pointconv,liu2019relation}. For example, PointConv~\cite{wu2019pointconv} generates convolution kernels for a local neighborhood by transforming relative positions (relative to local centers). On the other hand, RSCNN~\cite{liu2019relation} uses more features including relative positions, absolute positions, and euclidean distances to generate kernels so that kernels are explicitly 3D geometry-aware. PAConv~\cite{xu2021paconv} additionally learns a kernel bank from which convolution kernels are generated. Though effective, these methods do not explicitly consider the local structures characterized by edges which can describe the underlying surface more accurately.  

\paragraph{Transformer-based methods}
Inspired by the success of Transformers~\cite{vaswani2017attention} in the Natural Language Processing field, many studies have tried to construct transformer-based methods for point cloud data~\cite{liu2021group,qin2022geometric}. Here, we focus on transformers that are designed for point cloud segmentation. Since self-attention is essentially a set operator, it can be straightforwardly applied to point clouds. For example, PAT~\cite{yang2019modeling} replaces the original multi-head self-attention with parameter-efficient group shuffle attention so that self-attention can be applied to massive point cloud data. PointASNL~\cite{yan2020pointasnl}, on the other hand, uses self-attention for learning the global characteristics of point clouds. The obtained global features are then combined with local ones learned by point convolutions so that the resulting features are both aware of the global and local characteristics of point clouds. Point Transformer~\cite{zhao2021point} applies vector attention instead of original self-attention to the local subset of points to increase the expressiveness while reducing computation costs. Inspired by the success of BERT~\cite{devlin2019bert}, Point-BERT~\cite{yu2022point} pre-trains the transformer architecture by masked point modeling and demonstrates its good transferability of the learned representations. Stratified Transformer~\cite{lai2022stratified} achieves efficient modeling of long-range dependencies using the stratified strategy which samples nearby points densely and distant points sparsely. Recently, geodesic self-attention~\cite{ligeodesic2022} is proposed to capture the long-range dependencies by introducing a metric on the Riemannian manifold. Point Transformer V2~\cite{wu2022pointv2} improves the efficiency of Point Transformer~\cite{zhao2021point} by proposing efficient vector attention and lightweight partition--based pooling.     

\subsection{Edge-aware methods} To better model the local structure, edge-aware methods incorporate edge information into network design in various ways. As is described in Section~\ref{sec: introduction}, we regard methods that explicitly use the output of neural network functions that receive features of a center point and its neighbor as input as edge-aware methods. In general, there are three types of edge-aware methods. First, some methods perform convolution on edge features. The representative of such methods is EdgeConv~\cite{wang2019dynamic} which performs convolution on edge features to better model the local geometric structure. This idea is generally applicable and thus is adopted in numerous following works (e.g., \cite{li2019deepgcns,liu2020closer,xiang2021walk}). Despite that explicit modeling of edges brings performance improvements, the above methods are hardly interpretable, making users wonder how edges are modeled to improve performance.    
Second, some methods regard edge information as a measure of semantic distance. This idea uses edges as spatial weights that indicate the relationships between a center point and its neighbors. Moreover, it is often combined with the idea of self-attention~\cite{zhao2019pointweb,zhao2021point,xiu2021enhancing,wu2022pointv2} which converts edges into normalized weights. 
For example, Point Transformer~\cite{zhao2021point} replaces the global dot product in the original self-attention with local edge features for modeling more fine-grained geometric relationships. This idea is proven particularly useful in point cloud analysis which leads to state-of-the-art performance in segmentation tasks. However, forced normalization of edge information may undermine its usefulness because absolute values of edge features contain rich structural information like smoothness.    
On the other hand, edge information may be used as an extra source of supervision. For instance, \cite{jiang2019hierarchical} constructs a dedicated branch for learning edge features and pairs it with a loss function that encourages spatial consistency.

Compared to the above edge-aware methods, DU is theoretically guaranteed and empirically verified to perform edge enhancement and suppression. Furthermore, DU is designed to model edge features directly rather than converting them to normalized weights. In addition, DU realizes automatic edge modeling without additional supervision.

\subsection{Diffusion methods}
The diffusion methods model the smoothing of data (e.g., images) as diffusion processes. The pioneering work of such methods is anisotropic diffusion~\cite{perona1990scale} which is designed to smooth out irrelevant discontinuities while preserving significant edges, achieving edge-preserving smoothing. Specifically, anisotropic diffusion performs smoothing in a locally adaptive manner by considering the edge strength of a center point. Points with high edge strength are preserved while those with low edge strength are smoothed. An important result of \cite{perona1990scale} is that such a behavior can be explained theoretically, making the method interpretable. 
Therefore, anisotropic diffusion and its theoretical/empirical behavior are extensively studied in image processing and numerous extensions are proposed~\cite{black1998robust,weickert1998anisotropic,weickert1999coherence,brox2006nonlinear}. Furthermore, the success of this idea is also adapted to other research fields (e.g., computer graphics~\cite{desbrun1999implicit,clarenz2000anisotropic,bajaj2003anisotropic}). 

The idea of diffusion is also adopted in modern deep learning studies, in particular the development of various graph neural networks (GNNs). For example, DCNN~\cite{atwood2016diffusion} proposes a new architecture that learns diffusion-based representations for graph-structured data. GDC~\cite{gasteiger2019diffusion} similarly models the information propagation as generalized graph diffusion aiming at modeling long-range relationships. ADC~\cite{zhao2021adaptive} further improves the flexibility of GDC in choosing the neighborhood size by learning the optimal neighborhood size from data. \cite{chamberlain2021grand} analyzes the existing architectures that can be interpreted as information diffusion on graphs using tools from partial differential equations (PDEs). Furthermore, the notion of diffusion is also used in probabilistic point cloud generation or point cloud completion. Inspired by the diffusion process in thermodynamics in which particles diffuse from the original distribution to a noise distribution, \cite{luo2021diffusion} models the point cloud generation as learning reverse diffusion process which transforms a noise distribution to a desired shape. \cite{zhou20213d} further integrates the denoising diffusion models with the point-voxel representation. PDR~\cite{lyu2022a} provides a new paradigm to better capture point density distribution when performing completion.  

Despite there exist many applications of the diffusion equation, extending the diffusion equation for adaptive edge enhancement and suppression for 3D point cloud segmentation remains under-explored. 

\section{Method}
\label{sec:method}
In this section, we first explain the motivation for introducing explicit edge modeling. Then, we provide a concise background about the diffusion equation on which we build our DU. 
Then, we give the continuous definition of DU and provide a theoretical analysis to show how DU achieves edge enhancement and suppression learning.
We subsequently introduce the discretized DU which can be applied to discrete point clouds and easily implemented on modern machines. 
Lastly, we design DU-Net for point cloud segmentation. 

\subsection{Motivation}
\begin{figure}[t]
    \centering
    \includegraphics[width=\linewidth]{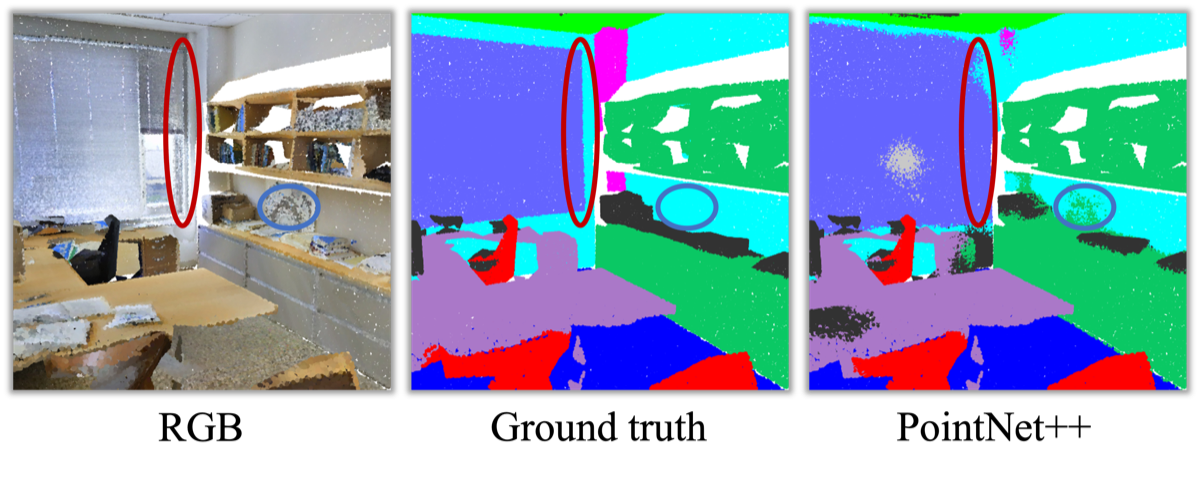}
    \caption{Examples of task and non-task edges. Red ellipses indicate task edges while blue ones indicate non-task edges.}
    \label{fig: edge_def}
\end{figure}

In this study, the term ``edge'' refers to one of the connections between two points. An edge indicates whether there is a change in attributes when moving from one point to another. For example, a large geometric change may indicate that the area is a corner. A change in color may also indicate the boundary of an object. Furthermore, these edges can be divided into two types, task edges and non-task edges, depending on the given task. Examples of such edges are shown in Fig.~\ref{fig: edge_def}. Task edges in tasks such as segmentation and edge detection are generally region boundaries, while non-task edges in these tasks are other discontinuities that do not correspond to region boundaries (e.g. noise or insignificant edges).

Point clouds naturally lack connectivity information, or edges. Therefore, by explicitly modeling edges, the network can learn to optimize the connectivity between points and thus better describe the local geometry. Explicit edge modeling is useful for tasks where connectivity between points is important. For example, in semantic segmentation, the network must distinguish between task and non-task edges to make high-fidelity predictions (see Fig.~\ref{fig: edge_def}), which is challenging because both low-level variations (e.g., geometric) and high-level semantics (e.g., object layout) must be considered simultaneously. To achieve this, feature representations of task edges should be enhanced/preserved, while non-task edges within region boundaries should be suppressed, since features of the same class should be similar to each other.

In this work, we accomplish explicit edge modeling in an \textit{interpretable} way using DUs that learn to enhance edges that are beneficial to the task and suppress edges that are irrelevant (Sec.~\ref{sec: theoretical_analysis}).

\subsection{Preliminary}
The diffusion equation describes the movement of diffusive substances from regions of higher concentration to lower concentration without creating or destroying mass~\cite{weickert1998anisotropic}. 
For example, when sugar is put into a glass of water, it diffuses evenly and sweetens the water.

Let ${u}({p}, t)$ denote the concentration at the position ${p}$ and time $t$. 
The amount of substances that flow through per unit area per unit time (the flux $s$) is described by Fick's law:
\begin{equation}
    {s} = - g \cdot \nabla {u} \;, 
    \label{eq: flux}
\end{equation}
where $\nabla$ denotes the gradient operator and $g$ denotes the diffusivity. This diffusivity denotes the rate of diffusion. The fact that diffusion processes do not create or destroy mass is expressed by the continuity equation:
\begin{equation}
    \partial_t {u} = - \mathrm{div}({s}) \;.
    \label{eq: continuity}
\end{equation}
The continuity equation indicates that the change of concentration over time is caused only by the flux, which is measured by the divergence operator ($\mathrm{div}$). Finally, the diffusion process is described by combining Eq.~\eqref{eq: flux} and Eq.~\eqref{eq: continuity}: 
\begin{equation}
    \partial_t {u} = \mathrm{div} (g \cdot \nabla {u}) \;\;\; t \ge 0\;,
    \label{eq: diffusion_equation}
\end{equation}
with the initial condition ${u}({p},0)={u}_0({p})$ and the boundary condition as appropriate.

\subsection{Continuous DU}
\label{sec: diffusion_unit}
Here we first adapt the general concept of diffusion to point cloud analysis. Informally, let us consider a case where $u(p,t)$ denotes a feature value or vector (e.g., sensor intensity or colors) of the position $p$ on a surface at time $t$. The strength of the edge is then described by $\nabla u$. Therefore, Eq.~\eqref{eq: diffusion_equation} describes the change of the feature at position $p$ in a short period of time (or in a discrete sense, the change of the feature after one step of diffusion). As can be seen, the behavior of the diffusion process is governed by the diffusivity $g$. $g$ is often a hand-crafted function of $u$, i.e., $g=g(u)$, and thus the diffusion process is adaptive to evolving feature values. This adaptability enables classic diffusion methods to realize edge-preserving filtering (e.g., \cite{perona1990scale} and \cite{black1998robust}). On the other hand, the behaviors of most existing choices of $g$ are theoretically restricted~\cite{weickert1998anisotropic}, that is, there is a limit to the total amount of smoothing or enhancement that can be performed by these methods.       

Extending the diffusion equation using existing choices of $g$ to our setting is nontrivial. First, $g$ is often a function with hand-crafted parameters that require considerable domain knowledge to determine. This is difficult in our setting due to the large variability of input point clouds. Second, adopting existing forms of $g$ may significantly limit expressiveness because of their theoretical restrictions. We want the method to be able to smooth and enhance features without restrictions if it is the right thing to do.     

We now propose a new way to extend the diffusion equation to our setting which is not affected by the above issues. To this end, we propose \textit{Diffusion Unit (DU)}, an algorithm that performs edge enhancement and suppression learning without restrictions.  
Formally, suppose a continuous spatial-temporal multi-channel point cloud $\mathbf{u} = \mathbf{u}(\mathbf{p}, t) = (u_1(\mathbf{p}, t), u_2(\mathbf{p}, t), ..., u_d(\mathbf{p}, t))$, where $d$ is the number of channels, $t$ is time, $\mathbf{p}$ denotes the position vector, and the initial condition is $\mathbf{u}(\mathbf{p}, 0) = \mathbf{h}$.
We define (continuous) DU as:
\begin{equation}
        \partial_t \mathbf{u} = \mathrm{div}\left(\Phi\left(\nabla \mathbf{u}\right)\right), \;\;\; t\ge0 \;,
    \label{eq: continuous_DU}
\end{equation} 
where $\nabla \mathbf{u} \in \mathbb{R}^{3 \times d}$ encodes channel-wise spatial (3D) gradient. We define $\Phi:\mathbb{R}^{3 \times d} \to \mathbb{R}^{3 \times d}$ as a trainable mapping of a form $\Phi(\nabla \textbf{u})=\left(\phi(\textbf{u}_x), \phi(\textbf{u}_y), \phi(\textbf{u}_z)\right)$ where $\phi:\mathbb{R}^d \to \mathbb{R}^d$. In other words, $\Phi$ applies $\phi$ to all $d$-dim components of the gradient $\nabla \textbf{u}$. In practice, we use an MLP to implement $\phi$.  

As can be seen, our idea is to replace the diffusivity weighted gradient with a trainable mapping that takes the gradient as input. In doing so, the equation is optimized with respect to data and no longer depends on hand-crafted parameters. Furthermore, the above form of $\Phi$ naturally makes the equation permutation-invariant when it is discretized, a useful property for point cloud analysis (this point will be explained in more detail in Section~\ref{sec:discretization}). In addition, the use of MLP enables the equation to learn and exploit channel correlations which are proven to be extremely helpful for vision tasks~\cite{hu2018squeeze}. Lastly, and most importantly, the design of $\Phi$ not only allows the equation to perform edge enhancement and suppression without restrictions but also enables us to understand the behavior of the equation in a mathematically transparent manner, making DU's behavior interpretable. This point will be fully discussed in the following section.         

\subsection{Theoretical analysis of edge enhancement and suppression learning}
\label{sec: theoretical_analysis}
\begin{figure}[t]
    \centering
    \includegraphics[width=0.7\linewidth]{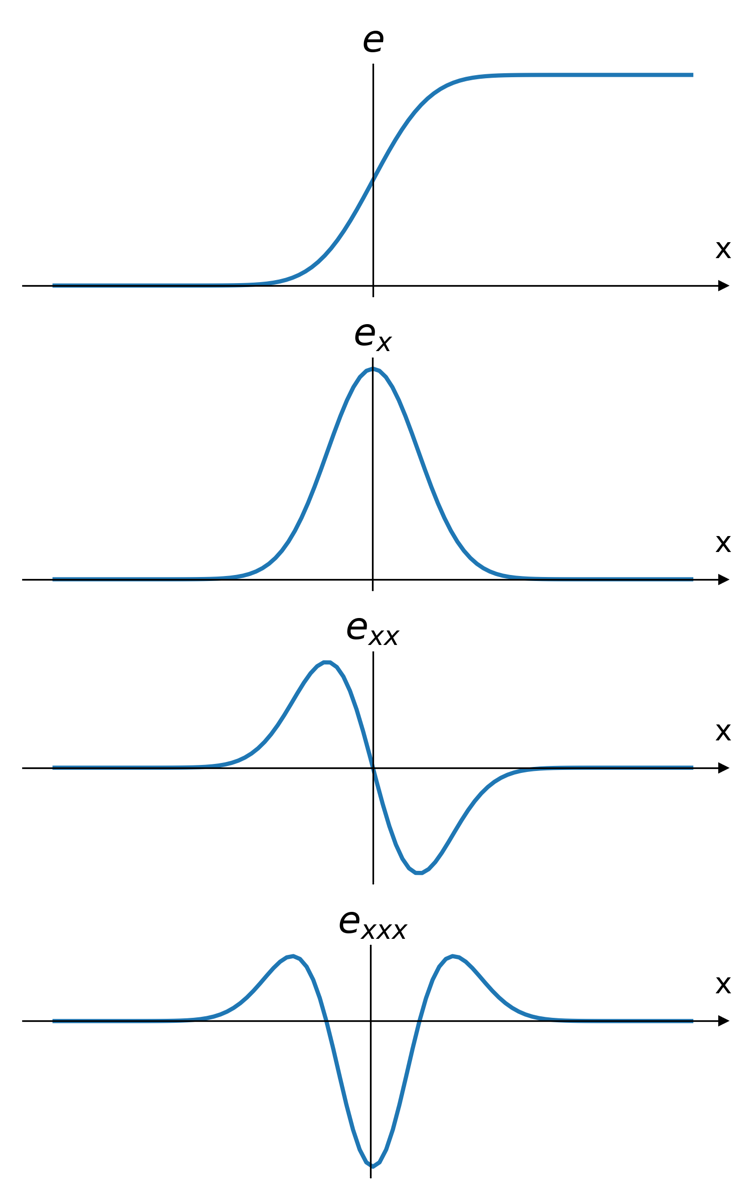}
    \caption{Profiles of the smoothed step edge (top) and its derivatives. At the inflection point, the second derivative is zero while the third derivative is smaller than 0.
    }
    \label{fig: edge_profile}
\end{figure}

In this part, we explain how DU performs edge enhancement and suppression learning.
For simplicity, we consider a step edge (i.e., an abrupt change in feature values) convolved by a Gaussian. Ideally, such a structure can be detected by the spatial gradient $\nabla \mathbf{u}$. Without loss of generality, we assume that the edge is aligned with $x$ axis ($\mathbf{u}_y=\textbf{u}_z=\textbf{0}$). The profile of a step edge and its derivatives are illustrated in Fig.~\ref{fig: edge_profile}. Now we focus on a single output channel $i$ of Eq.~\eqref{eq: continuous_DU} for brevity. In this setting, Eq.~\eqref{eq: continuous_DU} can be simplified as:
\begin{align}
    (\partial_t \textbf{u})_i
        &=
        \frac{\partial}{\partial x}\left(
            \phi_i\left(\textbf{u}_x\right)
            \right)
            =
            \nabla\phi_i \cdot \textbf{u}_{xx}\\
        &= \sum_{j=1}^{d} (\phi_i')_j \cdot(\textbf{u}_{xx})_j, \;\;\; i=1,2,...,d \;,
\end{align}
where $j$ indexes the input channel. We are interested in how the edge, i.e., $(\textbf{u}_x)_i$, changes during the diffusion process. In other words, we want to investigate the behavior of $(\partial_t \textbf{u}_x)_i$. This can be realized using the above equations: 
\begin{align}
        (\partial_t \textbf{u}_x)_i
        &= \left(
                \frac{\partial}{\partial t}\frac{\partial \textbf{u}}{\partial x}
            \right)_i
        =\frac{\partial}{\partial x}
                    \left(\frac{\partial \textbf{u}}{\partial t}\right)_i
                    \\
        &= \frac{\partial}{\partial x}\left(
            \sum_{j=1}^{d}\left(\phi_i'\right)_j
            \cdot \left(\textbf{u}_{xx}\right)_j
        \right) \\
        &= \sum_{j=1}^{d} (\phi_i^{''})_j \cdot (\textbf{u}_{xx})_j^2 + (\phi_i')_j \cdot (\textbf{u}_{xxx})_j \; .
        \label{eq: contribution}
\end{align}

As shown in Fig.~\ref{fig: edge_profile}, at the inflection point $(\textbf{u}_{xx})_j = 0$ and $(\textbf{u}_{xxx})_j < 0$. Therefore, the behavior of $(\partial_t \textbf{u}_x)_i$ is solely decided by the second term of Eq.~\eqref{eq: contribution}. Since $(\textbf{u}_{xxx})_j < 0$, the effect of a particular input channel $j$ to the output channel $i$ is determined by the sign of $(\phi_i')_j$. Specifically, channel $j$ has a positive impact if $(\phi_i')_j < 0$, whereas it has a negative impact if $(\phi_i')_j > 0$. Therefore, edge enhancement ($(\partial_t\textbf{u}_x)_i > 0$) or suppression ($(\partial_t\textbf{u}_x)_i < 0$) of a particular feature channel $i$ can be achieved by $\phi$ through summing up contributions from all input channels. Note that only task-beneficial edges are enhanced/suppressed because $\phi$ is trained w.r.t data and tasks.     

The empirical verification of edge enhancement and suppression behavior is presented in Section~\ref{sec: interpretation}. 

\subsection{Discrete DU}
\label{sec:discretization}
The discretization of Eq.~\eqref{eq: continuous_DU} is necessary because the 3D point clouds are discrete samples of continuous surfaces. Let $s$ and $n\in\mathcal{N}_s$ denote the center point and its neighbor, respectively. Then, $\textbf{u}_s, \textbf{u}_n \in \mathbb{R}^d$ denote the features of the center point and its neighbor. Eq~\eqref{eq: continuous_DU} is continuous both in time and space, so both parts must be discretized. For the time discretization, we adopt the simple explicit scheme~\cite{weickert1998anisotropic}: 
\begin{equation}
    \partial_t \textbf{u}_s    
    \approx \textbf{u}_s^{\tau+1} - \textbf{u}_s^\tau \;, 
    \label{eq: time_discretization}
\end{equation}
where $\tau$ denotes the discrete time index (iteration). 
For the space discretization, we use the following discretization scheme:
\begin{equation}
    \mathrm{div}\left(\Phi\left(\nabla \textbf{u}\right)\right)
    \approx \sum_{n \in \mathcal{N}_s} \phi\left(\textbf{u}_n^\tau - \textbf{u}_s^\tau \right)\;,
    \label{eq: spatial_discretization}
\end{equation}
where the gradient $\nabla u$ and divergence operator $\mathrm{div(\cdot)}$ are approximated by the element-wise difference $\textbf{d}_{sn}=\textbf{u}_n^\tau - \textbf{u}_s^\tau$ and summation $\sum_{n \in \mathcal{N}_s}\textbf{d}_{sn}$, respectively, similar to \cite{chamberlain2021grand}. In addition, this discretization scheme effectively makes DU an edge-aware method with the edge feature $\phi(\textbf{u}_n^\tau - \textbf{u}_s^\tau)$.   

Combining Eqs.~\eqref{eq: continuous_DU}, \eqref{eq: time_discretization}, \eqref{eq: spatial_discretization}, we have
\begin{equation}
    \textbf{u}_s^{\tau+1}
    =  \textbf{u}_s^\tau + 
    \sum_{n \in \mathcal{N}_s} \phi\left(\textbf{u}_n^\tau - \textbf{u}_s^\tau\right)\;.
    \label{eq: DU_discretization_pre}
\end{equation}

Eq.~\eqref{eq: DU_discretization_pre} has two important characteristics for deep learning--based point cloud analysis. First, the resulting equation belongs to the residual learning framework~\cite{he2016deep}, which makes optimization easier. Second, as we mentioned in Section~\ref{sec: diffusion_unit}, the equation is permutation-invariant because $\phi$ is a shared/pointwise MLP. This property is particularly useful in point cloud analysis, where the order of points is arbitrary and therefore the structured functions cannot be applied~\cite{qi2017pointnet}.               

Furthermore, we add some improvements from the perspective of engineering. First, we average the influence of neighbors in the second term of Eq.~\eqref{eq: DU_discretization_pre} as this slightly increases performance. We speculate that the averaging enhances the outlier robustness. 
Second, we apply Batch Normalization~\cite{ioffe2015batch} and ReLU~\cite{glorot2011deep} $\varphi = \mathrm{ReLU}\circ\mathrm{BatchNorm}$ to the second term to further facilitate optimization. 
As a result, the discretized DU is defined as:
\begin{equation}
    \textbf{u}_s^{\tau+1} 
    = \textbf{u}_s^\tau + 
        \varphi \left(\frac{1}{|\mathcal{N}_s|} \sum_{n \in \mathcal{N}_s} \phi\left(\textbf{u}_n^\tau - \textbf{u}_s^\tau\right)\right)\;.
    \label{eq: discretized_DU}
\end{equation}
The above definition of DU can be efficiently computed on modern computers and is used throughout this study. Its detailed computation flow is shown in Fig.~\ref{fig: arch} (middle).

\subsection{Network architecture of DU-Net}\label{sec: arch}
\begin{figure*}[t]
    \centering
    \includegraphics[width=.8\linewidth]{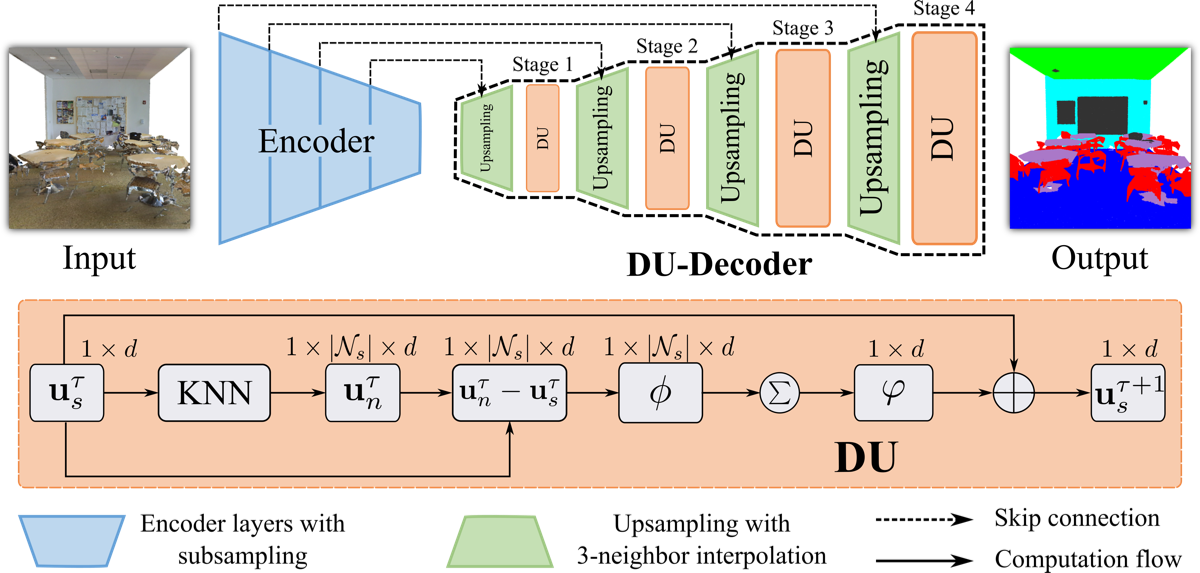}
    \caption{The network architecture used in this study and the computation flow of the discretized DU (Eq.~\eqref{eq: discretized_DU}). Top: the network architecture. Middle: computation flow of the DU. Here we assume the input is a point with a $d$-dimensional feature. $\sum$ and $\bigoplus$ denote neighborhood aggregation using summation and element-wise sum, respectively. }
    \label{fig: arch}
\end{figure*}

Using DU as a building block, we construct \textit{DU-Net} to tackle point cloud segmentation. An overview of the network architecture is shown in Fig.~\ref{fig: arch} (top). In designing DU-Net, we follow the most common encoder-decoder style to achieve efficient hierarchical learning. Such an architecture style is adopted by numerous previous works (e.g., \cite{qi2017pointnet++,thomas2019kpconv,xiang2021walk}) 

Similar to CNNs, the encoder is responsible for hierarchical feature abstraction. In each encoder stage, input point features are transformed and subsequently downsampled. The downsampling ensures an efficient encoding of multi-resolution characteristics, leading to a more discriminative feature representation. After downsampling, point features are passed to the next stage. We construct a CNN-like encoder based on KPConv~\cite{thomas2019kpconv}, which is a standard convolution operator adopted in many previous works for its excellent performance and ease of implementation~\cite{hu2020jsenet,lai2022stratified}. Specifically, we select the depthwise~\cite{sifre2014rigid} version to reduce the overall complexity. Following prior works~\cite{qi2017pointnet++,zhao2021point}, we adopt the farthest point sampling as the downsampling method.


The decoder recovers the original resolution of point clouds by performing successive upsampling followed by a DU. An upsampling layer receives the output of the previous layer and recovers the resolution of points in the adjacent upper stage of the encoder. We adopt U-Net~\cite{ronneberger2015u}--style skip connection to assist feature reconstruction. A DU is subsequently applied to the upsampled features so that task-beneficial edges of each resolution are kept sharpened while irrelevant discontinuities are smoothed out. 

We use DUs in the upsampling part because the upsampling part receives the features from the encoder part via skip connections (dotted arrows in Fig.~\ref{fig: arch}). This means that the input features of the DU are a direct combination of the encoder and decoder features. Therefore, it is considered sufficient to place the DU only in the upsampling part. In addition, we confirm experimentally that including DUs in the encoder does not improve performance.
We refer to our decoder as DU-decoder for brevity.

\section{Experiments}
\label{sec: exp}
In this section, we conduct experiments to answer the following questions:

\begin{enumerate}
\item Does DU really perform edge enhancement/suppression as described by the theoretical analysis in Section~\ref{sec: theoretical_analysis}?
\item Can DU-Net compete with recent cutting-edge networks?
\item Is the design choices of DU and DU-Net reasonable?
\end{enumerate}
To answer these questions, we use standard benchmarks of object part segmentation and scene segmentation tasks.


For object part segmentation, we use ShapeNet part dataset~\cite{yi2016scalable}, which contains 16,881 3D models that are classified into 16 object categories. Each model is annotated with several parts (less than 6) from 50 object part classes. For instance, an airplane category usually consists of parts including airplane body, wings, tails, and engine classes. 


For scene segmentation, we use Stanford large-scale 3D indoor spaces (S3DIS)~\cite{armeni20163d}. The dataset contains 3D scans taken from six areas including 272 rooms. For instance, rooms such as lobby, office, and conference rooms are included. Each point is annotated with a class from 13 categories. A room, for example, in the dataset includes classes such as floor, wall, door, and table.      
\subsection{Verifying the behaviors of DU}
\label{sec: interpretation}
\begin{figure}[t]
    \centering 
        \includegraphics[width=.95\linewidth]{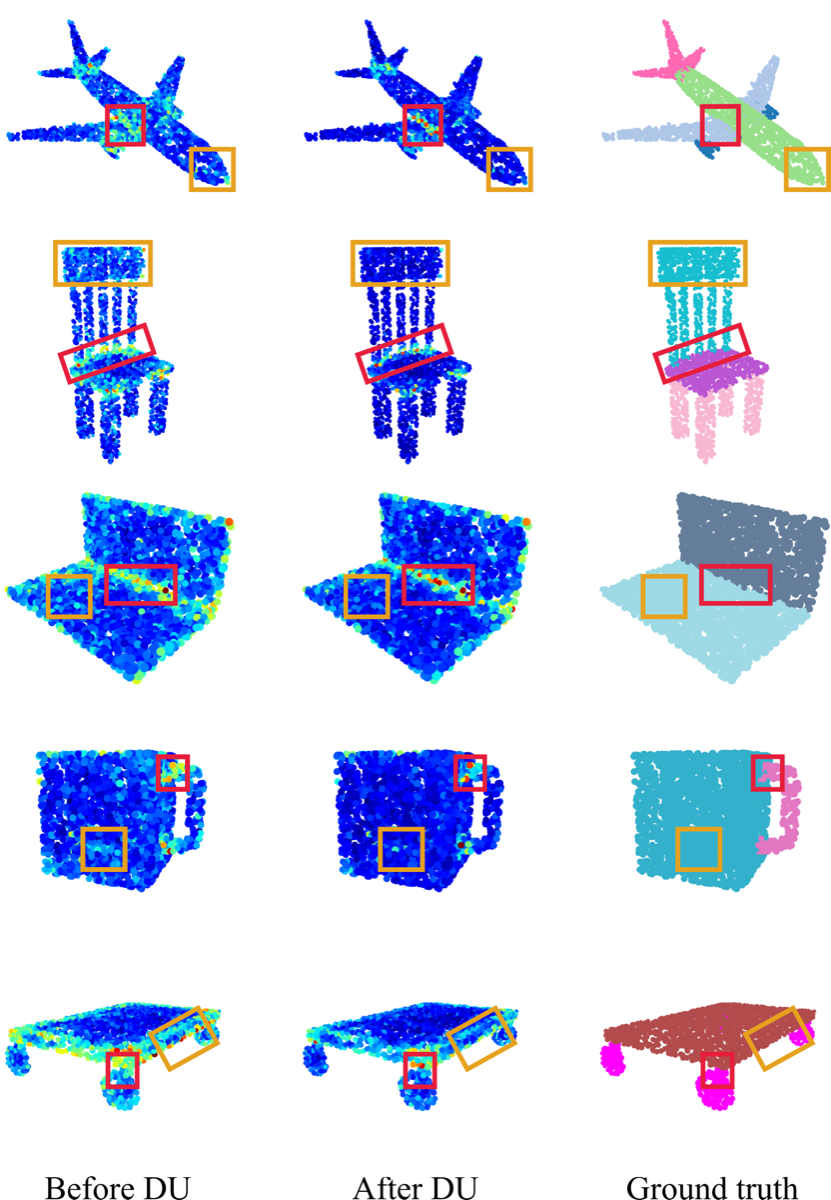}
    \caption{
    Examples of local feature smoothness before and after applying DU using the ShapeNet part test set. A low response indicates a smooth region while a high response indicates a non-smooth region. Red and orange rectangles indicate enhanced and suppressed edges after DU. The task-beneficial edges (in this case, part boundaries) are enhanced (become brighter) while within-part features are suppressed (become darker). 
    }
    \label{fig: behavior_shapenet}
\end{figure}

\begin{figure*}[t]
    \centering 
        \includegraphics[width=.9\linewidth]{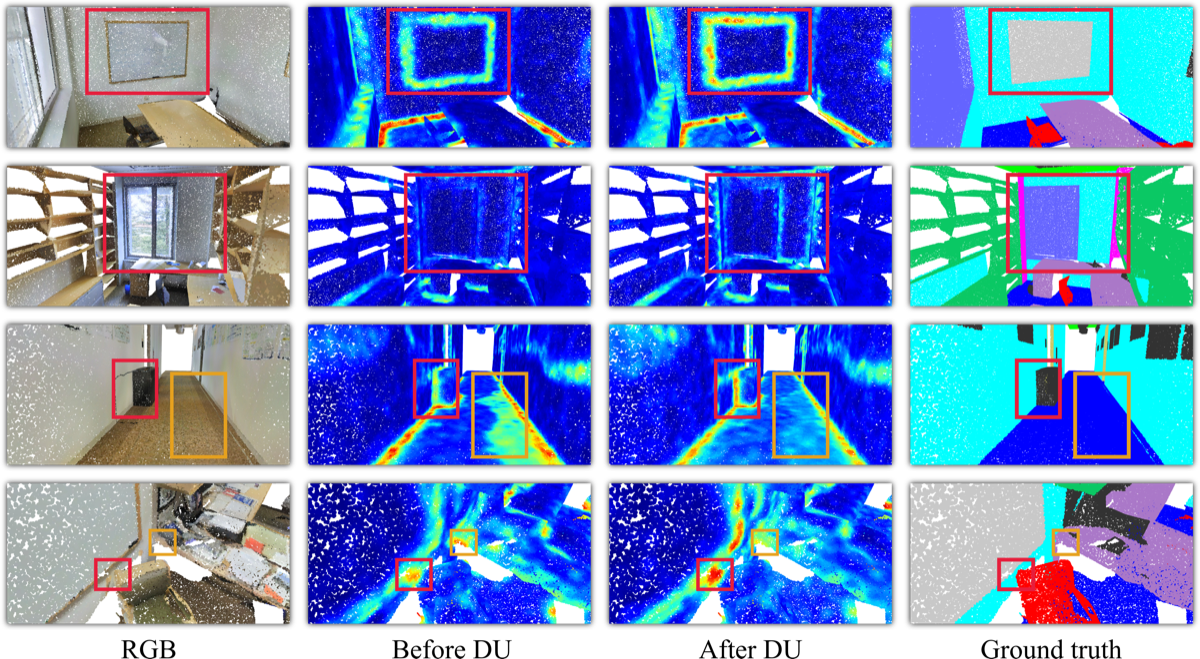}
    \caption{Examples of local feature smoothness before and after applying DU using the S3DIS Area 5 test set. A low response indicates a smooth region while a high response indicates a non-smooth region. Red and orange rectangles indicate enhanced and suppressed features after DU.
    The task-beneficial edges (in this case, object boundaries) are enhanced (become brighter) while intra-region features are suppressed (become darker). 
    }
    \label{fig: behavior_s3dis}
\end{figure*}
We have theoretically analyzed that DU learns to enhance or suppress edges in Section~\ref{sec: theoretical_analysis}. Here we empirically verify the edge enhancement and suppression behaviors of DU through visualizations. The features used for this experiment are taken from the last DU (stage 4 in Fig.~\ref{fig: arch}) since it is the closest layer to the final classification layer. To verify the effect of DU, we examine \textit{local feature smoothness} that summarizes how different the features are from their local neighbors. 
Formally, local feature smoothness is defined as
$||\sum_{n\in\mathcal{N}_s}\textbf{f}_n - \textbf{f}_s||$, where $\textbf{f}_s$ and $\textbf{f}_n$ denote the features of a center point and its neighbor from some layer. It can be observed that the smoothness value is closer to zero if the feature of the center point is similar to those of its neighbors while the value increases when they are dissimilar. Therefore, we can qualitatively grasp the edge enhancement/suppression behavior of DU by comparing smoothness values before and after applying DU.

Fig.~\ref{fig: behavior_shapenet} shows some examples of the change in local feature smoothness after applying DU from the ShapeNet part dataset. In this case, DU-Net is trained to assign part categories to each point. 
As can be seen, DU-Net is able to perform edge enhancement and suppression simultaneously. For instance, the third row shows a laptop whose boundary between its screen and keyboard is enhanced while the feature of its keyboards is smoothed. Therefore, DU successfully enhances the task-beneficial edges and suppressed the unhelpful discontinuities. 
Furthermore, we observe that DU can differentiate task-beneficial edges from other geometric edges (the top and last row). In the last row of Fig.~\ref{fig: behavior_shapenet}, for example, the sharp edges on the boundary (but not on the part boundary) of the table are aggressively suppressed after applying DU. On the other hand, the points near the part boundary are enhanced by DU. Thus, DU can understand the semantics of the objects and perform edge enhancement/suppression selectively.

Fig.~\ref{fig: behavior_s3dis} demonstrates the effect of DU in complex scenes. The top row shows that DU successfully locates and enhances the boundary of a whiteboard. As a result, the boundary becomes more salient. 
In the second row, DU also manages to enhance the object boundary (e.g., the boundary between the window and wall) that is barely distinguishable before applying DU. Interestingly, only the boundaries between different categories are enhanced while other discontinuities such as the textures within the window and the window frame remain suppressed; therefore, the DU clearly understands and distinguishes between task-beneficial and other edges. 
The third row shows a case in which edge enhancement and suppression occur simultaneously. The discontinuous edges that appear on the floor are smoothed after the DU while the boundary between a box and wall is enhanced. In the last row, DU precisely enhances the point where the wall and the chair are attached, while nicely suppressing the edge of the table that is close to the chair but not attached to it.  

Thus, the task-beneficial edge enhancement/suppression behavior of DU can be verified.

\subsection{Object part segmentation}
\label{sec: part_seg}

\begin{table}[t]
    \centering
    \caption{Result of object part segmentation (\%). The bold text indicates the best performance. * denotes methods that adopt voting for post-processing. Methods are arranged by year of publication.}
    \begin{tabular}{l|ccc}
        \toprule
        Method
        & Year
        & ImIoU
        & CmIoU
        \\
        \midrule
        PointNet~\cite{qi2017pointnet}
        & CVPR'17
        & 83.7
        & 80.4
        \\
        PointNet++~\cite{qi2017pointnet++}
        & NIPS'17
        &85.1 
        &81.9
        \\
        PointCNN~\cite{li2018pointcnn}
        & NIPS'18
        &86.1
        &84.6
        \\
        PointConv~\cite{wu2019pointconv}
        & CVPR'19
        &85.7
        &82.8
        \\
        SSCN*~\cite{graham20183d}
        & CVPR'18
        & 86.0
        & -
        \\
        RSCNN*~\cite{liu2019relation}
        & CVPR'19
        &86.2 
        &84.0
        \\
        KPConv*~\cite{thomas2019kpconv}
        & ICCV'19
        &86.4 
        & 85.1
        \\
        PAConv*~\cite{xu2021paconv}
        & CVPR'21
        &86.1 
        & 84.6
        \\
        Point Transformer~\cite{zhao2021point}
        & ICCV'21
        & 86.6
        & 83.7
        \\
        CurveNet*~\cite{xiang2021walk}
        & ICCV'21
        & 86.8
        & -
        \\
        Stratified Transformer*~\cite{lai2022stratified}
        & CVPR'22
        & 86.6
        & 85.1
        \\
        \midrule
        Ours (w/o DU)
        & -
        & 86.2
        & 83.6
        \\
        Ours (w/ DU)
        & - 
        & 86.7
        & 84.5
        \\
        Ours (w/ DU)*
        & -
        & \textbf{87.1}
        & \textbf{85.2}
        \\
        \bottomrule
    \end{tabular}
    \label{tab: result of object part segmentation}
\end{table}

For this experiment, we adopt the standard train-test split~\cite{qi2017pointnet++}. All available points are used for input. 3D coordinates along with surface normal information are used as the input features. We adopt the standard data augmentation strategy for point cloud analysis. Specifically, random anisotropic scaling in the range of [0.66, 1.5] and random translation in the range of [-0.2, 0.2] are used for data augmentation following prior works (e.g., \cite{wang2019dynamic}). We train the model using one NVIDIA Tesla V100 GPU. The model is trained for 150 epochs. SGD is used for optimization with an initial learning rate of 0.1. The learning rate is decayed by 0.1 when the epoch reaches 90 and 120. Following the common practice~\cite{thomas2019kpconv,liu2019relation,xu2021paconv}, we use voting for post-processing. We use the most widely adopted instance-wise mIoU (ImIoU) and category mIoU (CmIoU) defined in \cite{qi2017pointnet} for performance metrics. 
As for the loss function, we adopt a standard cross entropy loss.

The quantitative result is listed in Table~\ref{tab: result of object part segmentation}. DU-Net achieves state-of-the-art performance both in terms of ImIoU and CmIoU. This reveals that DU-Net not only performs well in the major classes but also achieves satisfactory accuracy for each class. Moreover, although the plain network (w/o DU in Table~\ref{tab: result of object part segmentation}) fails to compete with the cutting-edge networks, DU succeeds in improving its performance, achieving the best performance. 

\begin{figure}[t]
     \centering
    \includegraphics[width=0.85\linewidth]{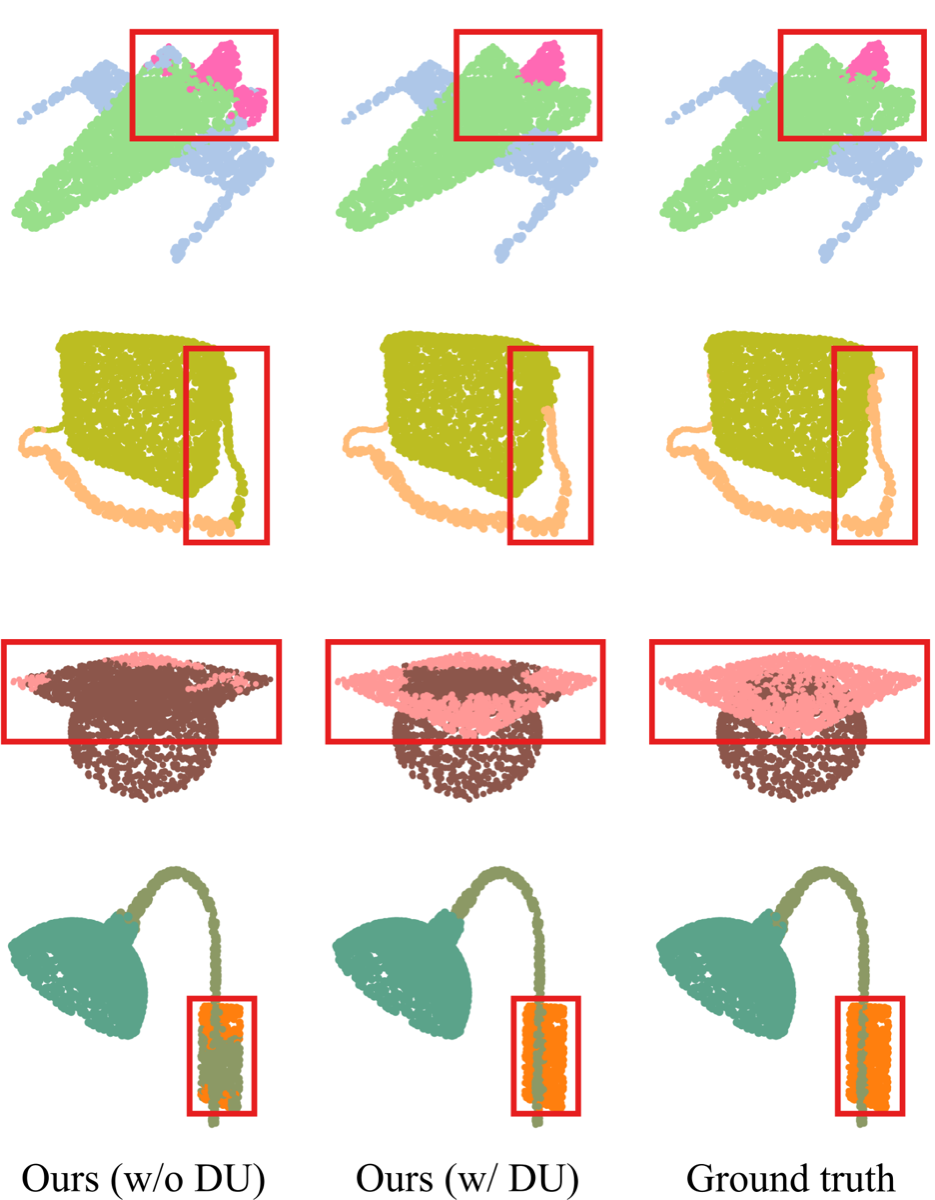}
    \caption{
        Qualitative results of object part segmentation. The red rectangles show the improvements brought by DU. As can be seen, DU encourages the model to produce smoother and more boundary-aware predictions.
    }
    \label{fig: qualitative results of object part segmentation}
\end{figure}

Next, we compare the predictions of DU-Net and the plain network to provide some insights into how DU improves performance. The results are shown in Fig.~\ref{fig: qualitative results of object part segmentation}. 
First, we find that DU-Net produces smoother predictions compared with the plain counterpart. For instance, the example of an airplane in the top row shows that the plain network has difficulty in distinguishing between the tail and body, resulting in ragged predictions. In contrast, DU-Net is able to produce smooth predictions even though the shape of the tail is similar to that of the body. We believe that the adaptability of DU in handling edges successfully suppresses the unhelpful edges, which in turn produces smoother predictions. 
Second, it is found that DU-Net tends to be more boundary-aware. In the second row, although both networks misclassify some of the points on the bag handle as bag body points, the result of DU-Net is much more accurate than the one of the plain counterpart, revealing that DU-Net has a better localization of the boundary between the bag body and handle. We conjecture that the ability to enhance task-beneficial edges makes features near boundaries more discriminative. As a result, DU-Net can detect part boundaries more precisely, thereby facilitating smooth predictions within boundaries and mitigating cross-boundary misclassifications (the third and last row). 


\subsection{Scene segmentation}
\label{sec: scene_seg}
\begin{table*}[t]
    \centering
    \caption{Result of scene segmentation (\%). The bold text indicates the best performance. Methods are arranged by year of publication.}
    \begin{tabular*}{\tblwidth}{L|L|C|CCCCCCCCCCCCC}
        \toprule
        Method
        & Year
        & mIoU
        & ceil.
        & floor
        & wall
        & beam
        & col.
        & wind.
        & door
        & chair
        & table
        & book.
        & sofa
        & board
        & clut.
        \\
        \midrule
        PointNet~\cite{qi2017pointnet}
        & CVPR'17
        & 41.1
        & 88.8
        & 97.3
        & 69.8
        & 0.5
        & 3.92
        & 46.3
        & 10.8
        & 52.6
        & 58.9
        & 40.3
        & 5.9
        & 26.4
        & 33.2
        \\

        SegCloud~\cite{tchapmi2017segcloud}
        & 3DV'17
        & 48.9
        & 90.1
        & 96.1
        & 69.9
        & 0.0
        & 18.4
        & 38.4
        & 23.1
        & 75.9
        & 70.4
        & 58.4
        & 40.9
        & 13.0
        & 41.6
        \\
        PointCNN~\cite{li2018pointcnn}
        & NIPS'18
        &57.3
        & 92.3
        & 98.2
        & 79.4
        & 0.0
        & 17.6
        & 22.8
        & 62.1
        & 80.6
        & 74.4
        & 66.7 
        & 31.7
        & 62.1
        & 56.7
        \\
        PointWeb~\cite{zhao2019pointweb}
        & CVPR'19
        & 60.3 
        & 92.0
        & 98.5
        & 79.4
        & 0.0
        & 21.1
        & 59.7
        & 34.8
        & 88.3
        & 76.3
        & 69.3
        & 46.9
        & 64.9
        & 52.5
        \\
        MinkowskitNet~\cite{choy20194d}
        & CVPR'19
        & 65.4
        & 91.8
        & 98.7
        & 86.2
        & 0.0
        & 34.1
        & 48.9
        & 62.4
        & 81.6
        & 89.8
        & 47.2
        & 74.9
        & 74.4
        & 58.6
        \\
        
        KPConv~\cite{thomas2019kpconv}
        & ICCV'19
        & 67.1
        & 92.8
        & 97.3
        & 82.4
        & 0.0
        & 23.9
        & 58.0
        & 69.0
        & 91.0
        & 81.5
        & 75.3
        & 75.4
        & 66.7
        & 58.9
        \\
        
        
        
        JSENet~\cite{hu2020jsenet}
        & ECCV'20
        & 67.7
        & 93.8
        & 97.0
        & 83.0
        & 0.0
        & 23.2
        & 61.3
        & 71.6
        & 89.9
        & 79.8
        & 75.6
        & 72.3
        & 72.7
        & 60.4
        \\
        CBL~\cite{tang2022contrastive}
        & CVPR'22
        & 69.4
        & 93.9
        & 98.4
        & 84.2
        & 0.0
        & 37.0 
        & 57.7
        & 71.9
        & 91.7
        & 81.8
        & 77.8
        & 75.6
        & 69.1
        & 62.9  
        \\
        Point Trans.~\cite{zhao2021point}
        & ICCV'21
        &70.4
        & 94.0
        & 98.5
        & 86.3
        & 0.0
        &38.0 
        &63.4
        &74.3
        & 89.1
        & 82.4 
        & 74.3
        & 80.2
        & 76.0
        & 59.3
        \\
        Stratified Trans.~\cite{lai2022stratified}
        & CVPR'22
        &72.0
        & 96.2 
        & 98.7
        & 85.6
        & 0.0
        & 46.1
        & 60.0
        & 76.8
        & 84.5
        & 92.6
        & 75.2
        & 77.8
        & 78.1
        & 64.0
        \\
        PointMixer~\cite{choe2022pointmixer}
        & ECCV'22
        & 71.4
        & 94.2 
        & 98.2
        & 86.0
        & -
        & 43.8
        & 62.1
        & 78.5
        & 82.2
        & 90.6
        & 79.8
        & 73.9
        & 78.5
        & 59.4
        \\
        PointNeXt~\cite{qian2022pointnext}
        & NeurIPS'22
        & 70.8
        &-&-&-&-&-&-&-&-&-&-&-&-&-
        \\
        Point Trans. V2~\cite{wu2022pointv2}
        & NeurIPS'22
        & 71.6
        &-&-&-&-&-&-&-&-&-&-&-&-&-
        \\
        \midrule
        Ours (w/o DU)
        & -
        & 69.7
        & 94.8
        & 98.5
        & 85.0
        & 0.0
        & 31.7
        & 58.6
        & 75.7
        & 80.8
        & 91.8
        & 76.1
        & 75.3
        & 79.5
        & 58.8
        \\
        Ours (w/ DU)
        & -
        & \textbf{72.2}
        & 95.6
        & 98.6
        & 85.2
        & 0.0
        & 40.0
        & 60.7
        & 82.7
        & 83.1
        & 90.8
        & 83.5
        & 78.5
        & 75.9
        & 64.1
        \\
        \bottomrule
    \end{tabular*}
    
    \label{tab: result of scene segmentation}
\end{table*}
Similar to \cite{tchapmi2017segcloud}, we advocate using Area five for testing and others for training. Following the common practice~\cite{thomas2019kpconv,zhao2021point,lan2019modeling}, we first grid-subsample each room with a grid size of 4cm for training. For testing, all points are evaluated. We form an input point cloud by taking at most 24,000 points from a room. We use 3D coordinates and color information as input features. Random vertical rotation, random anisotropic scaling in the range of [0.66, 1.5], random jitter (Gaussian noise with zero mean and 0.01 standard deviation), random color drop with a probability of 0.2, and random color auto-contrasting with a probability of 0.2 are used for data augmentation, following~\cite{zhao2021point,lai2022stratified}. Similar to the previous work~\cite{zhao2021point}, we train the model for 76,500 iterations. For this dataset, four NVIDIA Tesla V100 GPUs are used for training. For optimization, AdamW~\cite{loshchilov2018decoupled} algorithm is used with an initial learning rate of 0.01. The learning rate is decayed based on the cosine annealing schedule. We use mIoU to quantitatively assess the performance.
We use the cross entropy loss with label smoothing to train the model. The smoothing factor is set to 0.2.

The results are reported in Table~\ref{tab: result of scene segmentation}. DU-Net achieves state-of-the-art performance in terms of mIoU, which means DU-Net performs well in terms of average performance.  
We find that DU-Net performs especially well in classifying classes such as door and bookcase (book. in Table~\ref{tab: result of scene segmentation}). Since these two classes on often appear on the wall, one possible reason for the strong performance is that the edge enhancement capability of DU makes the boundary feature more discriminative, easing the detection of subtle but crucial boundaries.  
As can be seen, the plain network (w/o DU) fails to compete with the recent strong methods, while incorporating DUs achieves state-of-the-art performance. Therefore, the edge enhancement and suppression behavior of DU plays a key role in achieving good performance. 
As for the class-wise performance, we can verify that DUs successfully improve the performance by around 0.1--8.3, demonstrating the general usefulness of edge enhancement and suppression learning in scene segmentation.  
\begin{figure*}[t]
     \centering
    \includegraphics[width=.9\linewidth]{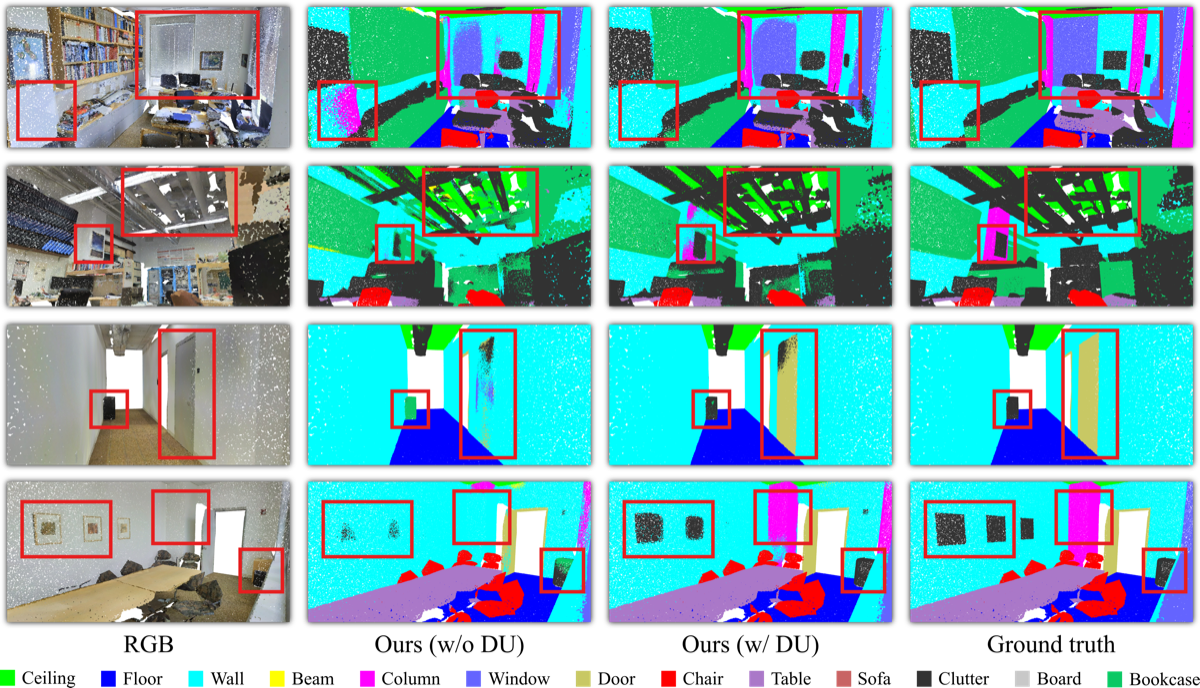}
    \caption{
    Qualitative results of scene segmentation using S3DIS. The red rectangles show the improvements brought by DU. In general, DU-Net (w/ DU) tends to better discriminate between similar classes (the first row) and produce smoother and boundary-aware predictions (the second row). In some cases, DU-Net correctly identifies objects that the plain counterpart (w/o DU) almost completely failed to recognize (the third and fourth rows).  
    }
    \label{fig: qualitative result of s3dis}
\end{figure*}

To intuitively understand how DU improves performance, we qualitatively analyze the effect of DU. The results are shown in Fig.~\ref{fig: qualitative result of s3dis}. We observe that DU-Net (w/ DU in Fig.~\ref{fig: qualitative result of s3dis}) generally better discriminates between similar classes compared with the plain counterpart (w/o DU in Fig.~\ref{fig: qualitative result of s3dis}). For instance, the example in the top row shows that the plain network is greatly confused with the window, wall, and column classes. DU-Net, on the other hand, correctly identifies those classes and produces much more accurate and smoother predictions. We believe that the adaptability of DU leads to enhanced sensitivity to object boundaries. Consequently, DU-Net can detect boundaries more accurately, which in turn facilitates spatial consistency within the detected boundaries.
Further, the example in the second row shows that DU makes the model more shape-aware. The plain network shows significant confusion in differentiating between pipes and the ceiling as the pipes are attached to the ceiling. In contrast, DU-Net shows much better predictions. We conjecture that explicit consideration of edges enables DU-Net to obtain more shape-aware representations, resulting in better discrimination.   
Moreover, we observe that DU-Net can make effective use of semantic information when geometric information is inadequate. In the third row, the plain net completely fails to identify the door likely because the door is embedded in the wall, making it geometrically difficult to differentiate the door from the wall. On the contrary, DU-Net succeeds in producing far better predictions. Similarly, in the last row, although the plain net manages to detect the centers of drawings, it fails to identify precise boundaries. In contrast, DU-Net produces fine predictions with accurate boundaries, demonstrating its advanced capability to utilize semantic information.    


\subsection{Design analysis}
\label{sec: design}

We further study the design of DU and DU-Net. All experiments are conducted on the S3DIS dataset and the experiment settings are the same as described in Section~\ref{sec: scene_seg}.

\begin{table}[t]
    \centering
    \caption{Ablation study on DU. Model 1 indicates the default design of DU. Numbers in parentheses denote the performance reduction compared with the default DU. All experiments are performed on the S3DIS dataset.}
    \begin{tabular*}{\tblwidth}{L|CCCC|L}
        \toprule
         Model
         & $|\mathcal{N}_s|$
         & Edge feat.
         & $\phi$
         & $\varphi$
         & mIoU
         \\
         \midrule
         1 (default)
         & 16
         & $\textbf{u}_n^\tau - \textbf{u}_s^\tau$
         & \checkmark
         & \checkmark
         & 72.2
         \\
         2
         & 16
         & $\textbf{u}_n^\tau - \textbf{u}_s^\tau$
         & -
         & \checkmark
         & 69.4 {\footnotesize \textcolor{BrickRed}{$(-2.8)$}}
         \\
         3
         & 16
         & $\textbf{u}_n^\tau - \textbf{u}_s^\tau$
         & \checkmark
         & -
         & 70.8 {\footnotesize \textcolor{BrickRed}{$(-1.4)$}}
         \\
         4
         & 16
         & $\textbf{u}_n^\tau - \textbf{u}_s^\tau$
         & - 
         & -
         & 69.8 {\footnotesize \textcolor{BrickRed}{$(-2.4)$}}
         \\
         5
         & 8
         & $\textbf{u}_n^\tau - \textbf{u}_s^\tau$
         & \checkmark
         & \checkmark
         & 70.9 {\footnotesize \textcolor{BrickRed}{$(-1.3)$}}
         \\
         6
         & 24
         & $\textbf{u}_n^\tau - \textbf{u}_s^\tau$
         & \checkmark
         & \checkmark
         & 71.2 {\footnotesize \textcolor{BrickRed}{$(-1.0)$}}
         \\
         7
         & 28
         & $\textbf{u}_n^\tau - \textbf{u}_s^\tau$
         & \checkmark
         & \checkmark
         & 70.6 {\footnotesize \textcolor{BrickRed}{$(-1.6)$}}
         \\
         8
         & 16
         & $\textbf{u}_n^\tau$
         & \checkmark
         & \checkmark
         & 69.4 {\footnotesize \textcolor{BrickRed}{$(-2.8)$}}
         \\
         \bottomrule
    \end{tabular*}
    \label{tab: ablation}
\end{table}

\paragraph{Ablation study on DU} The results are shown in Table~\ref{tab: ablation}. First, the effect of functions $\phi$ and $\varphi$ are investigated. 
The performance drops significantly by 2.8 when $\phi$ is removed. This is reasonable since the adaptability of DU is mostly provided by $\phi$. Therefore, the importance of $\phi$ in the design of DU is verified. 
Next, we observe that removing $\varphi$ from DU results in a reduction of 1.4. Recall that $\varphi=\mathrm{ReLU}\circ\mathrm{BatchNorm}$ is mainly applied to ease optimization. Thus, we believe that removing $\varphi$ increases the optimization difficulty, which adversely affects the performance. On the other hand, the reduction of the performance is less severe compared with removing $\phi$; hence, adaptability is relatively more important than optimization in the design of DU. 
As expected, removing both functions results in a significant reduction in performance. Interestingly, the reduction of performance caused by removing both functions is less severe than the one that removes $\phi$. We conjecture that the direct application of $\varphi$ without any transformation may lead to information loss since $\varphi$ contains $\mathrm{ReLU}$.        
Then, we vary $|\mathcal{N}_s|$ from 8 to 28 to analyze the effect of neighborhood size in the design of DU. As can be observed, the lower scores are obtained when $|\mathcal{N}_s|$ takes too small (8) or too large values (28). On the other hand, better performance is achieved when $|\mathcal{N}_s|$ takes intermediate values. Therefore, we set $|\mathcal{N}_s|$ to 16 throughout this study.   
Finally, we replace the neighbor difference in Eq.~\eqref{eq: discretized_DU} ($\textbf{u}_n^\tau - \textbf{u}_s^\tau$) with the point feature ($\textbf{u}_n^\tau$) to validate the edge feature we used. The point feature is often used in layers that perform neighbor aggregation (e.g., graph convolution layers). As a result, we observe a 2.8 decrease in mIoU when we replace the difference with the point feature. We speculate that the use of the point feature causes DU to perform neighborhood aggregation, thereby losing the ability to explicitly model edges.

\paragraph{Influence of DUs at different stages}
\begin{table}[t]
    \centering
    \caption{Influence of DUs at different stages. Stages indicate different resolutions (see Fig.~\ref{fig: arch}). The numbers of parameters (M) and floating-point operations (FLOPs) (G) are also reported. The numbers in parentheses denote the performance reduction compared with the default DU-Net (DU at all stages).}
    \begin{tabular*}{\tblwidth}{L|CC|L}
         \toprule
         Model
         & \#param. (M)
         & FLOPs (G)
         & mIoU  
         \\
         \midrule
         DU at all stages
         & 8.07
         & 11.94
         & 72.2
         \\
         DU at stage 1
         & 7.80
         & 10.05
         & 70.3 {\footnotesize \textcolor{BrickRed}{$(-1.9)$}}
         \\
         DU at stage 2
         & 7.18
         & 10.06
         & 70.6 {\footnotesize \textcolor{BrickRed}{$(-1.6)$}}
         \\
         DU at stage 3
         & 7.03
         & 10.07
         & 71.0 {\footnotesize \textcolor{BrickRed}{$(-1.2)$}}
         \\
         DU at stage 4
         & 6.99
         & 10.07
         & 70.3 {\footnotesize \textcolor{BrickRed}{$(-1.9)$}}
         \\
         w/o DU
         & 6.98
         & 9.43
         & 69.7 {\footnotesize \textcolor{BrickRed}{$(-2.5)$}}
         \\
         \bottomrule
    \end{tabular*}
    \label{tab: stage_influence}
\end{table}
Since DUs are applied to various stages of the decoder, we investigate the individual influence of DU at each stage in terms of performance and computational complexity. The results are listed in Table~\ref{tab: stage_influence}. From stages 1 to 4, the resolution increases because of upsamplings while the feature dimension decreases.  
First, it is found that a single DU can immediately improve performance. As can be seen, DU at each stage improves upon the plain network (w/o DU) by 0.6--1.3. Therefore, although the amount of improvement varies, each DU can make effective use of information at the corresponding stage. 
Second, we observe that the benefits brought by DU at each stage can be accumulated. Specifically, applying DUs at all stages achieves the best performance while the effect of a single DU is limited. We speculate that the adaptability of the DUs allows them to be tailored to each resolution, resulting in a cumulative improvement.  
Third, we find that because of the lightweight nature of the DUs, each DU produces improvement with only a slight increase in computation cost. For example, applying DU at stage 3 improves mIoU by 1.3, but only increases \#param. and FLOPs by 0.7\% and 6.7\%, respectively. Thus, an acceptable cost-performance tradeoff can be achieved when computational resources are limited.

\paragraph{Comparison with other methods} 
\begin{table}[t]
    \centering
    \caption{Comparison with other methods. Various methods and DU are compared by replacing DU with other methods in the decoder. All decoders are paired with the encoder of DU-Net. Type denotes the main operation in each decoder layer. The values in the parentheses indicate the performance reduction compared with DU-decoder.}    
    \begin{tabular*}{\tblwidth}{L|CL}
         \toprule
         Decoder layer
         & Type
         & mIoU  
         \\
         \midrule
         DU (Ours)
         & Edge-aware
         & 72.2
         \\
         Feature propagation~\cite{qi2017pointnet++}
         & MLP
         & 69.7 \textcolor{BrickRed}{$(-2.5)$}
         \\
         PointDeconv~\cite{wu2019pointconv}
         & Deconvolution
         & 70.0 \textcolor{BrickRed}{$(-2.2)$}
         \\
         Vector attention~\cite{zhao2021point}
         & Edge-aware
         & 70.3 \textcolor{BrickRed}{$(-1.9)$}
         \\
         \bottomrule
    \end{tabular*}
    \label{tab: different_decoder}
\end{table}
We compare DU with different methods to validate its effectiveness. Several methods of different types (second column in Table~\ref{tab: different_decoder}) are selected for comparison. Specifically, methods are compared with DU by replacing DU with those methods. The encoder of DU-Net is used as the default encoder. The results are listed in Table~\ref{tab: different_decoder}. We first compare DU with the feature propagation layer, which is a core layer of PointNet++~\cite{qi2017pointnet++}. This layer is a standard layer that has been adopted by numerous previous works (e.g., \cite{liu2019relation,thomas2019kpconv,xu2021paconv,lai2022stratified}). As can be seen, DU-decoder outperforms the feature propagation layer by 2.5, demonstrating its effectiveness.
The second method is PointDeconv, a deconvolution layer based on PointConv~\cite{wu2019pointconv}. We can see that DU-decoder outperforms pointDeconv by 2.2 mIoU. 
The third method is vector attention~\cite{zhao2021point}, which is the core component of Point Transformer~\cite{zhao2021point}. Vector attention uses edges as a similarity measure for generating attention weights. The result shows that using DU can provide a 1.9 mIoU gain compared with vector attention.

\paragraph{Effect of DU on voxel-based methods}
Voxel-based methods are another popular type of network for point cloud analysis. They use regular grids to approximate the connectivity between points; therefore, we expect that the explicit modeling of edges provided by DUs is also valid for those networks. To validate this point, we use two voxel-based methods, submanifold sparse convolutional networks (SSCN) and MinkowskiNet, as baselines. We reimplemented the two networks and conducted experiments according to our experimental setup. In addition, we also make sure that the reimplemented networks performed as well as or better than the original networks. Experiments related to SSCN and MinkowskiNet are performed using ShapeNet and S3DIS, respectively. The results are reported in Tab.~\ref{tab: effect on voxel-based methods}. 
As can be seen, DU improves SSCN by 0.3\% ImIoU and improves MinkowskiNet by 0.9\%. Consequently, we confirm that DU is applicable to voxel-based networks.

\begin{table}[t]
    \centering
    \caption{Effect of DU on voxel-based methods. We report ImIoU for SSCN and mIoU for MinkowskiNet. Numbers in parentheses indicate improvements brought by DU.}
    \begin{tabular*}{\tblwidth}{LLLL}
         \toprule
         Method
         & Dataset 
         & w/o DU
         & w/ DU
         \\
         \midrule
         SSCN~\cite{graham20183d}
         & ShapeNet
         & 86.1
         & 86.4 \textcolor{ForestGreen}{(+0.3)}
         \\
         MinkowskiNet~\cite{choy20194d}
         & S3DIS
         & 70.4
         & 71.3 \textcolor{ForestGreen}{(+0.9)}
         \\
         \bottomrule
    \end{tabular*}
    \label{tab: effect on voxel-based methods}
\end{table}

\section{Conclusion}
\label{sec:conclusion}
3D point clouds naturally lack structural information that describes the underlying continuous surfaces. To alleviate this issue, edge information has been used to describe the local structures of point clouds. While edges have proven useful, it is still unclear how they help to improve. In this study, we propose DU that allows us to handle edges in a principled and interpretable manner.
First, we theoretically figure out that DU performs task-beneficial edge enhancement and suppression. Second, we verify the result of theoretical analysis by intuitive visualizations. Third, we validate that the incorporation of DUs into the network design improves performance. 
DU-Net is constructed to tackle point cloud segmentation. Specifically, DU-Net achieves state-of-the-art performance in object part segmentation and scene segmentation. 
We believe that DU can provide users with a clear idea of how to use this method, what results to expect, and how to diagnose problems when it fails, which is valuable since most of the methods based on DNNs operate in a black-box manner. 

One possible limitation of DU is that it may fail when point clouds are significantly corrupted (e.g., by noise, density variations, etc.), since DU relies heavily on the spatial gradient of point features (Eq.~\eqref{eq: continuous_DU}). In addition, we suspect that DU may not be appropriate for tasks that do not benefit from explicit edge modeling. For example, unlike segmentation, DU may have a limited impact on classification tasks because optimizing point connectivity or edges is not closely related to the task objective.

From an application point of view, it will be interesting to apply DU to other domains where the modeling of edges is crucial. An example is damage detection using point clouds~\cite{xiu2020collapsed}. Damage is often characterized by non-smooth distributions; thus, it would be interesting to see how enhanced adaptability in handling edges influences the damage detection result.      

\section*{Acknowledgments}
This paper is based on results obtained from a project, JPNP20006, commissioned by the New Energy and Industrial Technology Development Organization (NEDO). We would also like to acknowledge the support from JSPS Grant-in-Aid for Scientific Research (21K12042).

\printcredits

\bibliographystyle{cas-model2-names}

\bibliography{cas-refs}




\end{document}